\documentclass[11pt]{article}

\usepackage[hyperref]{acl}

\usepackage{orcidlink} 
\usepackage{amsmath,amssymb,amsfonts}
\usepackage{amsthm}
\usepackage{mathrsfs}
\usepackage{xcolor}
\usepackage{textcomp}
\usepackage{booktabs}
\usepackage{algorithm}
\usepackage{algorithmicx}
\usepackage{algpseudocode}
\usepackage{listings}
\usepackage{tabularx}
\usepackage{graphicx}
\usepackage{multirow}
\usepackage{times}
\usepackage{latexsym}
\usepackage{microtype}

\usepackage{hyperref}

\theoremstyle{plain}

\theoremstyle{definition}

\title{Eyes All Around: Design and Analysis of 360\textdegree\ LiDAR Perception Using Equivariant Feature Learning in Unstructured Traffic}

\author{
Pranav Darshan\orcidlink{0009-0004-5586-3994}$^{1}$ \quad
Raghuveer Narayanan Rajesh\orcidlink{0009-0002-8511-4260}$^{1}$ \quad
M. Uttara Kumari\orcidlink{0000-0001-9628-4770}$^{2}$ \\[6pt]
$^{1}$Department of Computer Science and Engineering \\
RV College of Engineering, Bengaluru 560059, India \\
{\small \texttt{pranavdarshan.cs22@rvce.edu.in}} \\[4pt]
$^{2}$Department of Electronics and Communication Engineering \\
RV College of Engineering, Bengaluru 560059, India
}

\begin{document}
\maketitle

\begin{abstract}
Perception in dense, unstructured urban traffic remains a major challenge for autonomous driving because of the wide variety of road users, frequent occlusions, irregular motion patterns, and the lack of standardized road layouts. Although recent LiDAR based 3D object detectors have shown strong performance in structured driving scenarios, most are developed and evaluated for limited field of view settings, and their behavior under full surround 360\textdegree{} sensing is still not well understood. This paper studies a 360\textdegree{} LiDAR perception pipeline for autonomous driving, with particular attention to panoramic sensing, azimuthal sector wise spatial processing, and transformation equivariant feature extraction in complex urban scenes. The paper presents a practical 360\textdegree{} perception framework that combines sector wise panoramic processing with rotation equivariant sparse convolutions and evaluates its behavior on a custom Ouster OS0 LiDAR dataset collected across diverse Indian urban traffic conditions. The results show generally stable detection across several object classes, with the strongest performance for cars at 92.02/90.51, buses at 80.53/76.34, and trucks at 78.59/74.16, while lower scores for pedestrians at 67.45/61.02, cyclists at 73.21/69.54, and motorcyclists at 71.20/68.13 reflect the greater difficulty of detecting smaller and more variable road users in dense urban scenes.
\end{abstract}

\section{Introduction}
\label{sec:intro}

Reliable perception of the surrounding environment in urban traffic remains one of the central challenges in autonomous driving and advanced driver assistance systems (ADAS). The difficulty arises from dense interactions among diverse road users, frequent occlusions, and highly variable scene configurations \cite{bib1,bib2,bib3,bib4,bib5}. Achieving comprehensive situational awareness is essential for safety critical functions such as collision avoidance, blind spot monitoring, and adaptive maneuvering, especially in unstructured traffic that does not follow well marked lane based layouts \cite{bib6,bib7}.

Among the available sensing modalities, LiDAR has emerged as an important component because it captures detailed three dimensional scene structure with relatively little dependence on ambient lighting. This makes it well suited for detecting and localizing heterogeneous road users in complex environments \cite{bib8,bib9}. Recent advances in panoramic LiDAR sensors, including devices such as the Ouster OS0, have enabled full 360\textdegree{} coverage with dense point measurements and have expanded perception beyond the constraints of earlier limited field of view systems.

Despite this progress, most existing perception algorithms and benchmark datasets, many of which were developed around Velodyne sensors and structured environments such as KITTI, are still not well aligned with the demands of dense and unstructured urban traffic \cite{bib9,bib10,bib11}. In practice, such methods often struggle with heterogeneous road users, non standard vehicle types, and unpredictable motion patterns, all of which are common in Indian traffic scenes \cite{bib12,bib13,bib14}. Moving to 360\textdegree{} LiDAR sensing also introduces challenges that are not fully addressed in conventional forward facing pipelines. When the scene is observed from all directions, arbitrary viewpoint changes, azimuthal discontinuities, and object splitting near sector boundaries can affect detection consistency, making it difficult to directly extend narrow field of view perception strategies to panoramic sensing.

To address this gap, this work investigates a full surround LiDAR perception pipeline that combines sector wise point cloud partitioning with transformation equivariant feature extraction. The novelty of this work lies not in proposing a completely new detector, but in formulating and examining how an equivariance aware detection pipeline can be adapted to a panoramic 360\textdegree{} sensing setting for dense, unstructured urban traffic. This is important because full surround perception introduces geometric and spatial challenges that are much less prominent in conventional forward looking setups. By organizing the point cloud into azimuthal sectors and using equivariant feature representations, the proposed framework makes it possible to study how detection behavior changes with viewing direction, vehicle heading, and boundary induced fragmentation. In this sense, the work offers a system level contribution by showing how sector aware panoramic processing and equivariant learning can be brought together in a practically usable perception pipeline for real urban traffic scenes. Building on prior work \cite{bib15,bib16} that demonstrated the value of equivariant representations in more controlled settings, this work extends that idea to the richer geometric context of full surround LiDAR perception.

For evaluation, a 360\textdegree{} LiDAR dataset was collected and annotated using an Ouster OS0 sensor across a range of Indian urban environments. The dataset captures varying traffic densities, occlusion patterns, and scene geometries, and includes six road user classes: pedestrians, cars, cyclists, motorcyclists, trucks, and buses. All training and quantitative analysis in this work are performed using data collected under a single and consistent 360\textdegree{} OS0 sensing configuration, which helps reduce cross dataset inconsistencies and allows the work to remain grounded in a realistic urban deployment setting.

The primary contributions of this work are summarized as follows:
\begin{itemize}
\item A full-surround 360\textdegree{} LiDAR perception framework designed for dense and unstructured urban traffic environments with heterogeneous road users and complex scene layouts.
\item A sector-wise panoramic processing strategy combined with transformation-equivariant feature extraction to improve robustness under arbitrary viewing directions and azimuthal discontinuities.
\item An overlap-aware spatial decomposition approach for reducing object fragmentation and improving detection consistency near sector boundaries in panoramic LiDAR scenes.
\item Evaluation on a 360\textdegree{} Ouster OS0 LiDAR dataset representing realistic Indian urban traffic conditions under a consistent sensing configuration.
\end{itemize}

\section{Related Works}
\label{sec:related}

LiDAR-based 3D object detection has advanced quickly with the introduction of deep learning architectures that operate directly on raw point clouds. Early foundational works such as PointNet \cite{bib8} and PointNet++ \cite{bib17} showed that neural networks can handle unordered point sets effectively without relying on hand-crafted geometric features. Follow-up methods, including DGCNN \cite{bib18}, KPConv \cite{bib19}, and PVCNN \cite{bib20}, introduced richer local neighborhood modeling and hierarchical feature extraction, leading to stronger performance on point cloud classification and segmentation tasks.

Beyond architecture design, domain adaptation methods such as Complete \& Label \cite{bib21} address distribution shifts in LiDAR point clouds, particularly for semantic segmentation when models are transferred across datasets. This line of work emphasizes the need to handle changes in data distribution, which is also a key consideration for perception in diverse urban environments.

While point-based approaches offer considerable flexibility, their computational cost has motivated voxel-based and hybrid representations for large-scale outdoor scenes. Methods such as VoxelNet \cite{bib22}, SECOND \cite{bib23}, and PointPillars \cite{bib24} discretize 3D space to enable efficient convolutional processing. Hybrid detectors, including PV-RCNN \cite{bib25} and CenterPoint \cite{bib26}, combine voxel-level context with point-level refinement, while sparse convolution frameworks \cite{bib27} and cylindrical coordinate representations \cite{bib28} further improve scalability for large-scale perception.

Despite their strong performance on established benchmarks, these detectors are typically designed for limited field-of-view LiDAR configurations and are mostly evaluated in structured traffic environments. Their behavior under full-surround sensing and highly unstructured traffic conditions is less well explored.

\subsection{Datasets for Autonomous Driving and Unstructured Environments}
\label{subsec:datasets}

Most widely used 3D detection models are developed and evaluated on structured driving datasets such as KITTI \cite{bib9}, which employ forward-facing LiDAR sensors with a restricted field of view. As a result, many architectures are implicitly optimized for narrow-view sensing and lane-structured traffic.

To broaden environmental diversity and sensing coverage, datasets such as nuScenes \cite{bib10}, Waymo Open Dataset \cite{bib11}, Lyft \cite{bib29}, ApolloScape \cite{bib30}, PandaSet \cite{bib31}, and A2D2 \cite{bib32} provide large-scale, multi-sensor data across varied environments. However, these datasets predominantly represent Western urban conditions with relatively homogeneous traffic patterns and well-regulated road infrastructure, and are typically curated under sensor layouts and evaluation protocols tailored to structured driving scenarios.

In contrast, datasets focusing on unstructured traffic---such as IDD \cite{bib12}, IDD-3D \cite{bib13}, and DriveIndia \cite{bib14}---capture dense and heterogeneous interactions among road users that are characteristic of many emerging economies. DriveIndia mainly provides RGB imagery for 2D perception, whereas IDD-3D additionally includes synchronized LiDAR data suitable for 3D perception. In the present work, these datasets are used to contextualize perception challenges in unstructured traffic; all training and quantitative evaluation are conducted exclusively under a single, consistent 360\textdegree\ LiDAR sensor configuration to ensure fair internal comparison and avoid cross-dataset bias.

\subsection{Equivariant and Panoramic LiDAR Perception}
\label{subsec:equivariant}

Transformation-equivariant learning has been studied as a way to improve robustness to sensor rotations and viewpoint changes. Group-equivariant convolutional networks \cite{bib16,bib33,bib34} formalize equivariance with respect to discrete or continuous transformation groups, enabling feature representations that change in a predictable way when the input is transformed. Building on these ideas, the Transformation Equivariant Detector (TED) \cite{bib15} incorporates equivariant feature extraction into 3D object detection pipelines, achieving more consistent predictions under rotated or perturbed viewpoints in limited field-of-view settings.

Related approaches such as SE(3)-Transformers \cite{bib35}, SpinNet \cite{bib36}, and LieConv \cite{bib37} extend equivariance to broader classes of continuous geometric transformations. While these methods offer strong theoretical guarantees, they are usually evaluated under constrained sensing configurations and do not explicitly analyze how equivariance interacts with panoramic LiDAR sensing or sector-based spatial partitioning. Equivariant graph-based models have also been explored for modeling physical dynamics, as in spatio-temporal attentive GNNs \cite{bib38}, highlighting the broader applicability of equivariant representations beyond static perception.

In contrast, the present work examines equivariant feature extraction under full 360\textdegree\ LiDAR sensing by coupling group-equivariant kernels with sector-aware spatial decomposition. This setting enables analysis of rotational consistency and boundary-induced object fragmentation, effects that frequently arise in panoramic LiDAR perception but are not explicitly addressed in narrow field-of-view detection pipelines.

Panoramic perception has received increasing attention as autonomous systems move toward surround-view sensing to improve situational awareness. Beltr\'{a}n et al.\ \cite{bib39} demonstrate the benefits of combining 360\textdegree\ LiDAR with multi-camera inputs for comprehensive scene understanding. Bird's-eye-view (BEV)-centric fusion frameworks such as BEVFusion \cite{bib40}, HDNet \cite{bib41}, and related approaches \cite{bib42,bib43,bib44} integrate information from multiple sensors into a shared spatial representation, enabling robust multi-modal reasoning. Recent attention-based BEV methods, such as SWA-SOP \cite{bib45}, incorporate spatially aware windowed attention for semantic occupancy prediction in autonomous driving. Bidirectional fusion strategies such as BiCo-Fusion \cite{bib46} exploit complementary LiDAR and camera cues for semantic- and spatial-aware 3D detection.

Despite their effectiveness, many of these methods rely on precise camera--LiDAR calibration and are designed for specific sensor layouts and operating conditions. Adapting such frameworks directly to modern panoramic LiDAR platforms, including sensors such as the Ouster OS0, is challenging due to differences in angular resolution, encoding schemes, coverage patterns, and the increased prevalence of occlusions and boundary effects under full-surround sensing.

\section{Dataset and Preprocessing}
\label{sec:dataset}

This section describes the construction of the dataset, the preprocessing steps used to maintain annotation consistency, and the experimental scope adopted in this work. The objective was to evaluate 3D detection under a single, consistent \(360^\circ\) LiDAR sensing configuration representative of dense and unstructured Indian urban traffic, while retaining compatibility with established detection pipelines.

\subsection{Dataset and Evaluation Protocol}
\label{subsec:dataset_protocol}

The dataset used in this work was developed in collaboration with the Wipro IISc Research and Innovation Network (WIRIN) Centre of Excellence for Autonomous Vehicles, Bengaluru. Data acquisition was performed using an Ouster OS0 LiDAR sensor, which provides full \(360^\circ\) horizontal coverage, dense spatial sampling, and \(16\)-bit intensity measurements. To capture the operational characteristics of unstructured Indian traffic, recordings were conducted in dense urban environments such as multi-lane intersections, arterial corridors, and mixed-use streets populated by diverse road users, including cars, trucks, buses, motorcycles, cyclists, pedestrians, and auto-rickshaws.

In total, \(5{,}200\) LiDAR frames were collected and manually annotated for \(3\)D object detection. Each frame contains \(3\)D bounding-box annotations for six object classes: car, pedestrian, cyclist, motorcyclist, truck, and bus. Across the complete annotated set, the dataset comprises approximately \(36{,}000\) cars, \(15{,}000\) motorcyclists, \(8{,}000\) pedestrians, \(5{,}200\) trucks, \(4{,}800\) cyclists, and \(3{,}000\) buses. This distribution reflects the density and heterogeneity typical of Indian urban traffic and provides a challenging evaluation setting for panoramic LiDAR perception.

All quantitative results reported in this work are obtained exclusively on this OS0-based dataset. To ensure a fair and leakage-resistant evaluation protocol, the data were partitioned into disjoint training and test splits using a \(70/30\) division at the sequence level. Frames belonging to the same continuous recording sequence were assigned entirely to either the training set or the test set, thereby reducing temporal correlation between development and evaluation samples.

\subsection{Preprocessing and Training Considerations}
\label{subsec:preprocessing}

Although all model training and evaluation were carried out solely on the OS0 dataset, limited use of external data was necessary during preprocessing design. Statistics from approximately \(7{,}000\) frames of the KITTI dataset~\cite{bib9} were analyzed to verify preprocessing conventions, validate annotation handling, and derive reasonable object size priors for anchor configuration. Their role was restricted to preprocessing validation and implementation of sanity checks. As a result, all reported performance comparisons remain confined to a single sensing domain under a uniform \(360^\circ\) LiDAR configuration.

An important preprocessing step arose from the fact that the WIRIN OS0 dataset and KITTI follow different coordinate conventions and bounding box reference definitions. To enable reuse of established tooling without compromising label consistency, WIRIN style annotations were transformed into KITTI compatible coordinates prior to export. For a \(3\)D bounding box with height \(h\), where \((x_{\mathrm{WIRIN}}, y_{\mathrm{WIRIN}}, z_{\mathrm{WIRIN}})\) denote the original bounding box center coordinates in the WIRIN reference frame and \((x_{\mathrm{KITTI}}, y_{\mathrm{KITTI}}, z_{\mathrm{KITTI}})\) denote the transformed coordinates in the KITTI reference frame, the transformed center coordinates are defined as
\begin{multline}
(x_{\mathrm{KITTI}}, y_{\mathrm{KITTI}}, z_{\mathrm{KITTI}}) \\
= (-y_{\mathrm{WIRIN}},\,-z_{\mathrm{WIRIN}} + h/2,\,x_{\mathrm{WIRIN}})
\label{eq:coord_transform}
\end{multline}
where \(h\) represents the height of the \(3\)D bounding box. This transformation aligns the annotations with KITTI's convention of referencing boxes at ground contact. The yaw angle transformation is defined as
\begin{equation}
\theta_{\mathrm{KITTI}} = -\theta_{\mathrm{WIRIN}}
\label{eq:yaw_transform}
\end{equation}
where \(\theta_{\mathrm{WIRIN}}\) and \(\theta_{\mathrm{KITTI}}\) represent the object orientation angles in the WIRIN and KITTI coordinate systems, respectively. The transformed labels were then exported in KITTI compatible text format. The correctness of this mapping was verified by projecting the transformed \(3\)D boxes into the corresponding \(2\)D image plane and visually inspecting the alignment between point clouds, annotations, and camera images.

During training, point clouds were augmented using random azimuthal rotations to expose the model to a broader range of sensor orientations. For each scene, yaw perturbations were sampled within a predefined range, for example within \(\pm 120^\circ\). In the present pipeline, however, rotational robustness is not attributed to augmentation alone. The primary mechanism is the transformation-equivariant design of the backbone, while data augmentation serves as a complementary optimization strategy.

Because full surround LiDAR introduces failure modes that are less apparent in narrow field-of-view benchmarks, each \(360^\circ\) point cloud was further partitioned into overlapping angular sectors. The overlap was introduced as an engineering control to ensure that objects straddling azimuthal boundaries remained fully represented in at least one sector, thereby reducing fragmentation at sector edges. In addition, an optional temporal fusion component was considered for qualitative analysis of occlusion effects in crowded scenes. In this setting, features extracted from a fixed window of consecutive LiDAR frames were fused through concatenation followed by lightweight channel wise weighting. However, all primary quantitative results reported in this paper were obtained using single frame inference, and temporal fusion is included only for completeness and future analysis.

This work is intended as an offline analysis of detection behavior under panoramic sensing and unstructured urban traffic conditions rather than as a real time deployment implementation. Accordingly, inference latency, hardware-specific optimization, asynchronous execution, and system-level runtime constraints are outside the scope of the present work.

The OS0 dataset serves as an internal evaluation platform rather than a public benchmark. Owing to contractual and confidentiality constraints associated with industrial collaboration, the dataset cannot currently be released publicly. This limitation also constrained large-scale re-implementation of external baseline detectors on the same data under identical conditions. Consequently, the empirical analysis in this work focuses on method behavior, robustness trends, and system-level design considerations under a single, controlled sensing configuration.

\begin{figure*}[t]
\centering
\includegraphics[width=\textwidth]{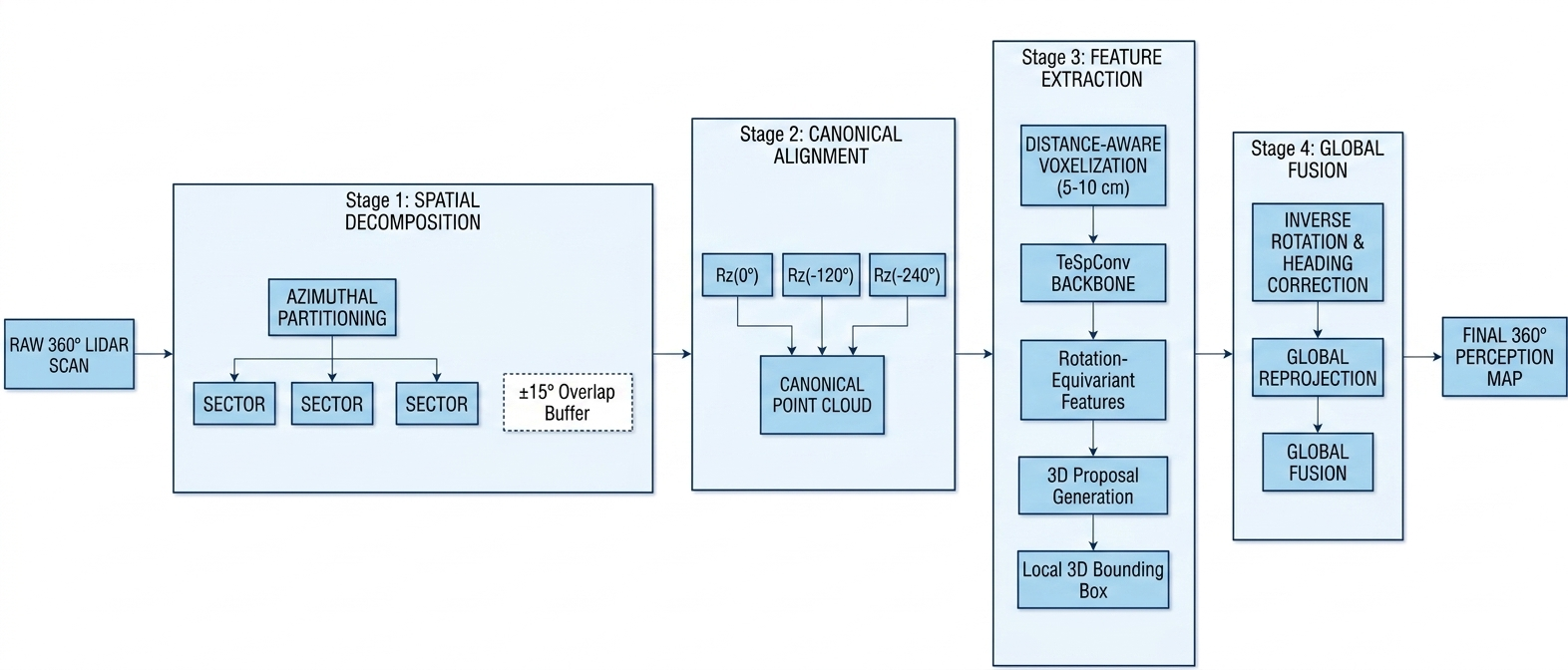}
\caption{The proposed split--normalize--merge perception pipeline.}
\label{fig:pipeline}
\end{figure*}

\begin{figure*}[t]
\centering
\includegraphics[width=\textwidth]{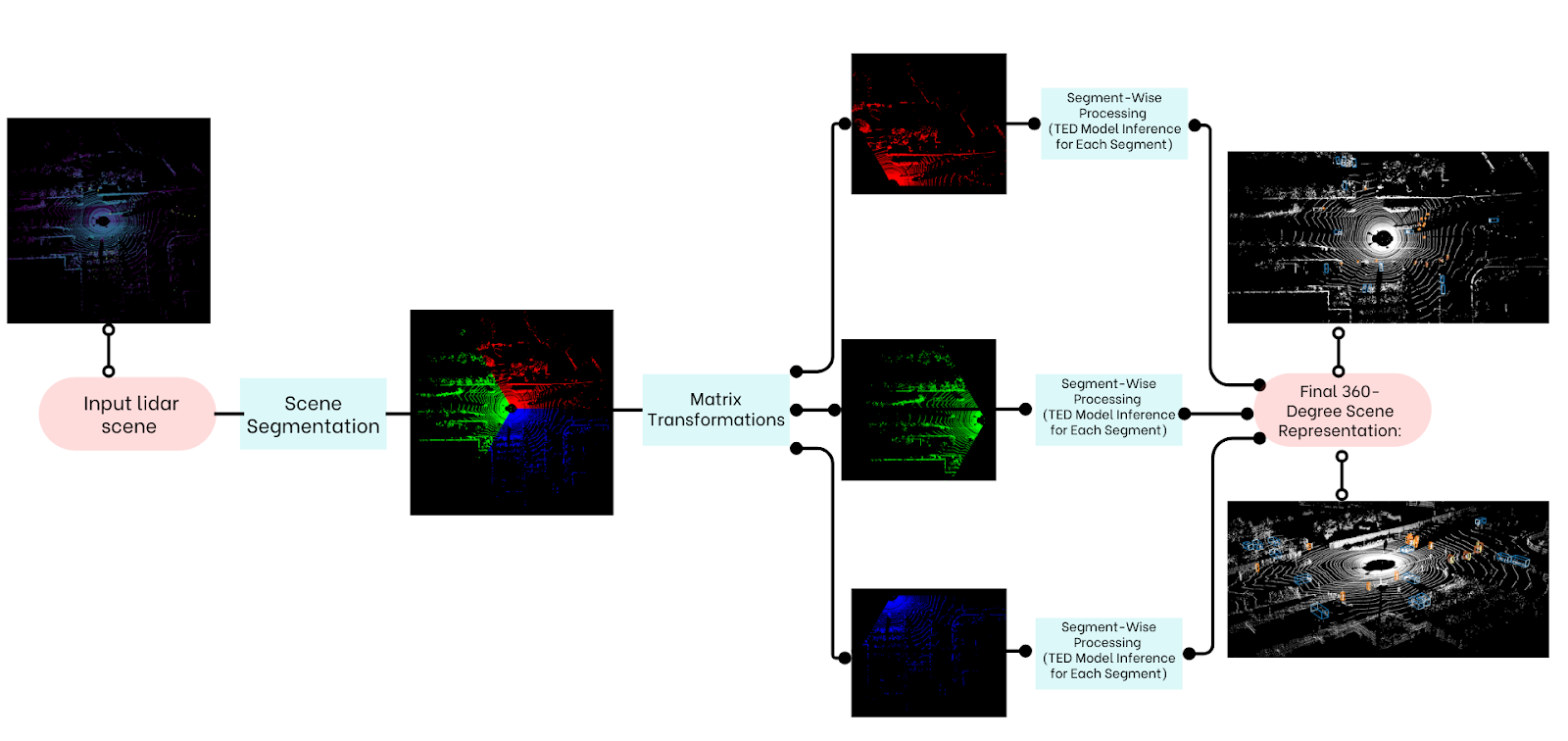}
\caption{Overview of the studied 360\textdegree{} LiDAR perception pipeline. Stages (A)--(D) denote azimuthal sector partitioning, equivariant feature extraction, temporal feature fusion, and full-surround 3D object detection, respectively.}
\label{fig:overview}
\end{figure*}

\section{Implementation of 360\textdegree\ LiDAR Perception}
\label{sec:implementation}

This section describes the perception pipeline used in the work and the design choices made to support analysis of full-surround LiDAR sensing in unstructured urban traffic. The pipeline combines sector-wise processing, distance-aware voxelization, and transformation-equivariant feature extraction. Although the implementation is tailored to the characteristics of the Ouster OS0 sensor, the design principles are applicable to other panoramic LiDAR configurations.

As illustrated in Figures~\ref{fig:pipeline} and~\ref{fig:overview}, the proposed perception pipeline unfolds across four interconnected stages. First, raw 360\textdegree{} LiDAR scans are ingested from the sensor. These scans are then partitioned azimuthally into overlapping sectors, a design choice that proves critical for maintaining detection continuity across the full surround view. Within each sector, spatial features are extracted and, when temporal data is available, aggregate information across consecutive frames to capture motion cues. Finally, the sector-level predictions are fused back together, yielding a unified detection output that spans the entire environment.

This staged approach directly tackles several persistent challenges in panoramic LiDAR perception. Object fragmentation, where a single vehicle or pedestrian gets split between adjacent sectors, is mitigated through overlapping sector strategy. The pipeline also exhibits reduced sensitivity to vehicle heading changes, since features are computed in a rotationally consistent manner. Perhaps most importantly, temporal aggregation helps the system maintain stable detections even during short-term occlusions, which frequently occur in dense urban traffic when one object briefly obscures another.

\subsection{Spatial Decomposition}
\label{subsec:spatial}

Each 360\textdegree\ LiDAR scan is divided into three overlapping azimuthal sectors, with each sector spanning 120\textdegree\ in the horizontal plane. This partitioning allows inference to be performed within localized angular windows, which reduces sensitivity to clutter and abrupt viewpoint changes commonly observed in dense urban scenes. Adjacent sectors overlap by 15\textdegree\ on either side to ensure that objects near sector boundaries are fully represented in at least one sector. Sector assignment is determined using the azimuth angle $\theta$ of each point $p \in \mathcal{P}$, measured in the horizontal plane relative to the sensor frame, where $\mathcal{P}$ denotes the full point cloud. The $k$-th sector $S_k$ is defined as:
\begin{equation}
S_k = \bigl\{ p \in \mathcal{P} \mid \theta \in [\theta_k^-, \theta_k^+] \bigr\},
\quad k \in \{0,1,2\},
\label{eq:sector}
\end{equation}
where $\theta_k^- = k\cdot\tfrac{2}{3}\pi - \tfrac{1}{12}\pi$ and
$\theta_k^+ = (k{+}1)\cdot\tfrac{2}{3}\pi + \tfrac{1}{12}\pi$,
and the three sectors together cover the full 360\textdegree\ scan. Overlap is treated as a practical control mechanism rather than a novel algorithmic contribution. Duplicate detections introduced by overlapping regions are resolved at a later stage using standard non-maximum suppression.

Prior work has shown that adapting voxel resolution to local point density can improve robustness to non-uniform LiDAR sampling \cite{bib47,bib48}. Motivated by this observation, the proposed mechanism adopts an adaptive voxelization strategy tailored to sector-wise panoramic LiDAR geometry. Point density in panoramic LiDAR scans decreases with increasing distance from the sensor, making uniform voxel resolution inefficient. To account for this, distance-aware voxelization is used, where voxel size depends on the radial distance $d$ of each point. Specifically, a resolution of 5~cm is used for points within 20~m of the sensor, while a coarser resolution of 10~cm is used for points between 20~m and 50~m. Points beyond 50~m are optionally downsampled depending on the setting. These thresholds were selected empirically to preserve geometric detail in near-field regions while limiting memory usage and computation in mid-range areas. The resulting sparse voxel grids form the input to the 3D backbone and provide stable training behavior across the evaluated scenes.

Raw intensity values produced by the OS0 sensor differ in scale from those used in legacy datasets such as KITTI. To obtain a consistent input representation, a simple per-scan intensity normalization is applied:
\begin{equation}
I_\text{norm} = 255 \times \frac{I_\text{OS0} - I_\text{min}}{I_\text{max} - I_\text{min}},
\label{eq:intensity}
\end{equation}
where $I_\text{OS0}$ denotes the raw intensity measurement and $I_\text{min}$ and $I_\text{max}$ are the minimum and maximum intensity values within the scan. This mapping preserves relative reflectance patterns while producing values that are compatible with common LiDAR detection architectures.

\subsection{Class-Specific Anchor Design}
\label{subsec:anchor}

Anchor dimensions are derived directly from the OS0 dataset to better reflect the geometry of road users encountered in dense urban traffic. For each object class---car, pedestrian, cyclist, motorcyclist, truck, and bus---the dimensions of all annotated bounding boxes are collected, and empirical means and variances are computed. Anchors are then placed at the class mean and at selected offsets around the mean to capture intra-class variation, for example differences between small cars and larger sedans or between minibuses and full-size buses. This design improves anchor coverage for geometrically diverse classes and reduces bias toward vehicle dimensions typical of structured highway datasets.

\subsection{Transformation-Equivariant Feature Extraction}
\label{subsec:equivariant_feat}

To reduce sensitivity to vehicle heading and sensor orientation, the detection backbone incorporates transformation-equivariant feature extraction, following the principles introduced in the Transformation Equivariant Detector (TED) framework \cite{bib15}. Let $x$ denote the input point cloud representation and let $f(\cdot)$ represent the learned feature extraction function. A feature extractor is said to be equivariant if applying a geometric transformation $T_g$ to the input produces a corresponding and predictable transformation in the output feature space:
\begin{equation}
f(T_g(x)) = T_g(f(x)),
\label{eq:equivariance}
\end{equation}
where $T_g$ denotes a transformation associated with an element $g$ of the relevant subgroup of the special Euclidean group SE(3), which models rigid-body rotations and translations in 3D space. In this work, the dominant transformations correspond to rotations around the vertical axis of the LiDAR sensor.

By embedding rotational symmetry directly into the convolutional kernels, the pipeline reduces variance in feature space caused by viewpoint shifts. This follows the basic idea of geometric deep learning: encoding known symmetries into the network architecture lowers the burden on data augmentation.

In the context of unstructured traffic, where heterogeneous vehicles move in non-parallel directions and interact at unpredictable angles, this consistency helps ensure that the geometric signature of an object is preserved even as its relative orientation to the sensor changes across the 360\textdegree\ field of view. In this work, equivariance is enforced with respect to planar rotations around the vertical axis. For a rotation by angle $\theta$, where $\theta$ denotes the azimuthal rotation angle in the horizontal plane, the point coordinates $(x,y)$ are transformed into rotated coordinates $(x',y')$ as:
\begin{equation}
\begin{pmatrix} x' \\ y' \end{pmatrix} =
\begin{pmatrix}
\cos\theta & -\sin\theta \\
\sin\theta & \cos\theta
\end{pmatrix}
\begin{pmatrix} x \\ y \end{pmatrix},
\label{eq:rotation}
\end{equation}
where $(x,y)$ and $(x',y')$ represent the original and rotated planar coordinates, respectively.

In practice, voxelized sectors are processed using a sparse 3D convolutional backbone whose kernel structure and feature channels are organized to respect this rotational symmetry. Rotating the input induces a predictable transformation in feature space, which reduces the need for extensive rotation augmentation. This aspect is particularly important for panoramic sensing: while standard voxel-based models often rely on brute-force augmentation to ``learn'' rotations, the equivariant design makes the model inherently aware of the azimuthal continuity of the 360\textdegree\ scan. Unlike prior TED implementations that operate on a single narrow field of view, this work examines equivariant feature extraction in combination with sector-wise processing and subsequent global aggregation, with the aim of maintaining detection stability across all azimuthal boundaries.

To make the relationship between this work and the original TED framework explicit, Table~\ref{tab:ted_comparison} summarizes the main differences in sensor configuration, objectives, and evaluation setting between TED \cite{bib15} and the present pipeline.

\begin{table}[t]
\centering
\small
\caption{Comparison of the Setting with the Original TED Framework}
\label{tab:ted_comparison}
\begin{tabular}{@{}p{2.0cm}p{2.5cm}p{2.5cm}@{}}
\toprule
\textbf{Aspect} & \textbf{TED \cite{bib15}} & \textbf{Proposed Mechanism} \\
\midrule
Sensor field of view & Forward-facing LiDAR (limited azimuth) & Full 360\textdegree\ LiDAR sensing \\[4pt]
Input representation & Single continuous point cloud & Sector-wise partitioned panoramic point cloud \\[4pt]
Primary objective & Rotation robustness under narrow FOV & Analysis of equivariant behavior under panoramic sensing \\[4pt]
Equivariance type & Rotation equivariance around vertical axis & Rotation equivariance examined across azimuthal sectors \\[4pt]
Handling of boundary effects & Not explicitly addressed & Analyzed via overlapping sector partitioning \\[4pt]
Temporal modeling & Single-frame inference & Single-frame inference (optional temporal analysis) \\
\bottomrule
\end{tabular}
\end{table}

\subsection{Global Fusion}
\label{subsec:fusion}

Short-term occlusions are common in dense urban traffic, particularly for pedestrians and two-wheelers. To analyze the effect of temporal context, an optional feature-level aggregation mechanism is included. For each sector, features from a fixed number of consecutive scans are concatenated along the channel dimension and passed through a lightweight channel-wise weighting module. Temporal aggregation is disabled during standard evaluation and is used only for additional robustness analysis. Unless explicitly stated otherwise, all reported quantitative results correspond to single-frame inference.

After per-sector processing, detections are transformed back into the global coordinate frame and merged. Overlapping predictions across sector boundaries are resolved using non-maximum suppression in bird's-eye view. The resulting set of 3D bounding boxes provides a full-surround representation of the scene that can be used for downstream analysis and evaluation.

\section{Results and Analysis}
\label{sec:results}

A central challenge in evaluating 360\textdegree\ LiDAR perception is the limited direct comparability with detectors optimized for narrow field-of-view (FOV) datasets such as KITTI. Many standard 3D detection architectures are designed around front-facing coordinate systems and partial azimuthal coverage, making direct numerical comparison under panoramic sensing inherently biased. For this reason, the evaluation in this work focuses on internal ablation studies to analyze how sector-based partitioning and equivariant feature extraction address geometric challenges specific to full-surround perception in unstructured urban traffic.

This section presents quantitative and qualitative results obtained using the OS0-based 360\textdegree\ LiDAR dataset described in Section~\ref{sec:dataset}. The results are intended to analyze detection behavior under panoramic sensing and unstructured traffic conditions, rather than to establish numerical superiority over detectors evaluated on different datasets or sensing setups.

Detection performance is evaluated using standard 3D object detection metrics, including average precision (AP) computed both in bird's-eye view (BEV) and in full 3D space. A fixed IoU threshold of 0.7 is used for all object classes to keep the evaluation consistent and directly comparable. Metrics are reported separately for each of the six road-user categories: car, pedestrian, cyclist, motorcyclist, truck, and bus.

All reported results are based on single-frame inference unless explicitly noted otherwise. External datasets are not used for evaluation, and no results from other benchmarks are compared directly, since differences in sensor field of view and traffic characteristics would make such comparisons difficult to interpret.

Table~\ref{tab:results} summarizes the detection performance achieved on the OS0 dataset across the six object classes. Cars achieved the highest detection accuracy with a BEV AP of 92.02\% and a 3D AP of 90.51\%, reflecting the relatively large size and stable geometric structure of vehicles in panoramic LiDAR scans. Larger vehicle classes such as buses and trucks also produced strong performance, with buses reaching 80.53\% BEV AP and 76.34\% 3D AP, while trucks achieved 78.59\% and 74.16\%, respectively. In contrast, smaller and more dynamically varying categories remained more challenging. Cyclists obtained 73.21\% BEV AP and 69.54\% 3D AP, motorcyclists achieved 71.20\% and 68.13\%, and pedestrians produced the lowest results at 67.45\% BEV AP and 61.02\% 3D AP, mainly due to sparse point coverage, frequent occlusions, and high appearance variability in dense urban scenes.

The reported AP values should be interpreted in the context of the internal OS0 dataset and the controlled evaluation setup. They reflect detection behavior under full-surround sensing in dense urban traffic, rather than direct comparisons with detectors evaluated on narrow field-of-view benchmarks.

\begin{table}[t]
\centering
\caption{Average Precision (AP) in BEV and full 3D space for six object classes evaluated on the OS0-based 360\textdegree\ LiDAR dataset.}
\label{tab:results}
\begin{tabular}{@{}lcc@{}}
\toprule
\textbf{Class} & \textbf{BEV AP (\%)} & \textbf{3D AP (\%)} \\
\midrule
Car          & 92.02 & 90.51 \\
Pedestrian   & 67.45 & 61.02 \\
Cyclist      & 73.21 & 69.54 \\
Truck        & 78.59 & 74.16 \\
Bus          & 80.53 & 76.34 \\
Motorcyclist & 71.20 & 68.13 \\
\bottomrule
\end{tabular}
\end{table}

\begin{figure}[t]
\centering
\includegraphics[width=\columnwidth]{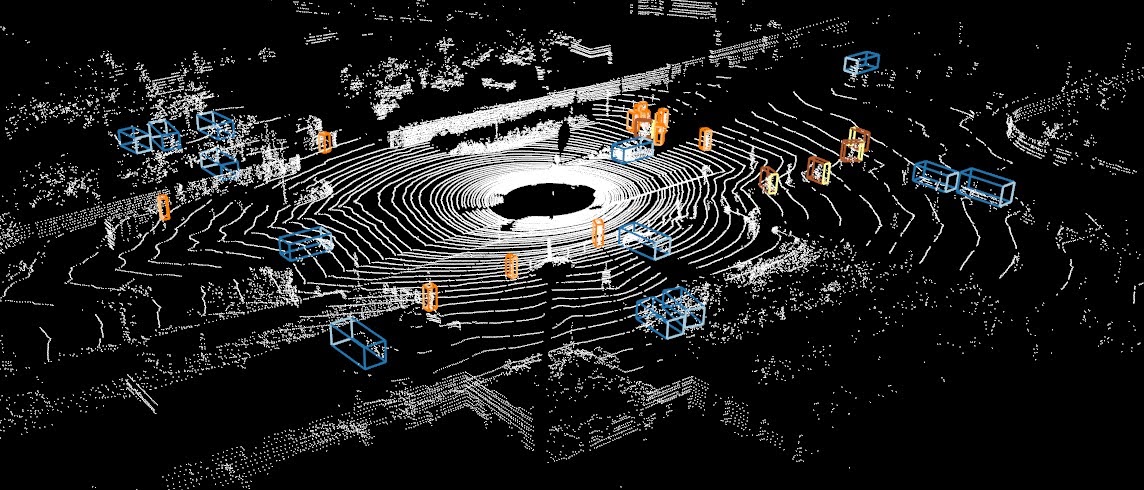}
\caption{Example 360\textdegree\ 3D detection results on the OS0 dataset using the studied pipeline. Predicted bounding boxes are color-coded by object class to illustrate performance across heterogeneous road users.}
\label{fig:detections}
\end{figure}

Figure~\ref{fig:detections} shows a few representative examples from dense urban scenes. In these frames, the pipeline is able to pick up a mix of nearby road users, including cars, two-wheelers, and pedestrians, often interacting in close quarters. When sector overlap is enabled, objects that fall near sector boundaries are still localized consistently, which helps to avoid the fragmentation artifacts that are otherwise common in full-surround scans.

Most of the visible failure cases relate to strong occlusions, very sparse point coverage for far-away objects, or confusing class boundaries between similar categories such as cyclists and motorcyclists. These issues are closely tied to the basic limits of LiDAR-based perception in unstructured traffic, rather than being quirks of this particular dataset.

\subsection{Effect of Sector-Based Processing}
\label{subsec:sector_effect}

\begin{figure}[t]
\centering
\includegraphics[width=\columnwidth]{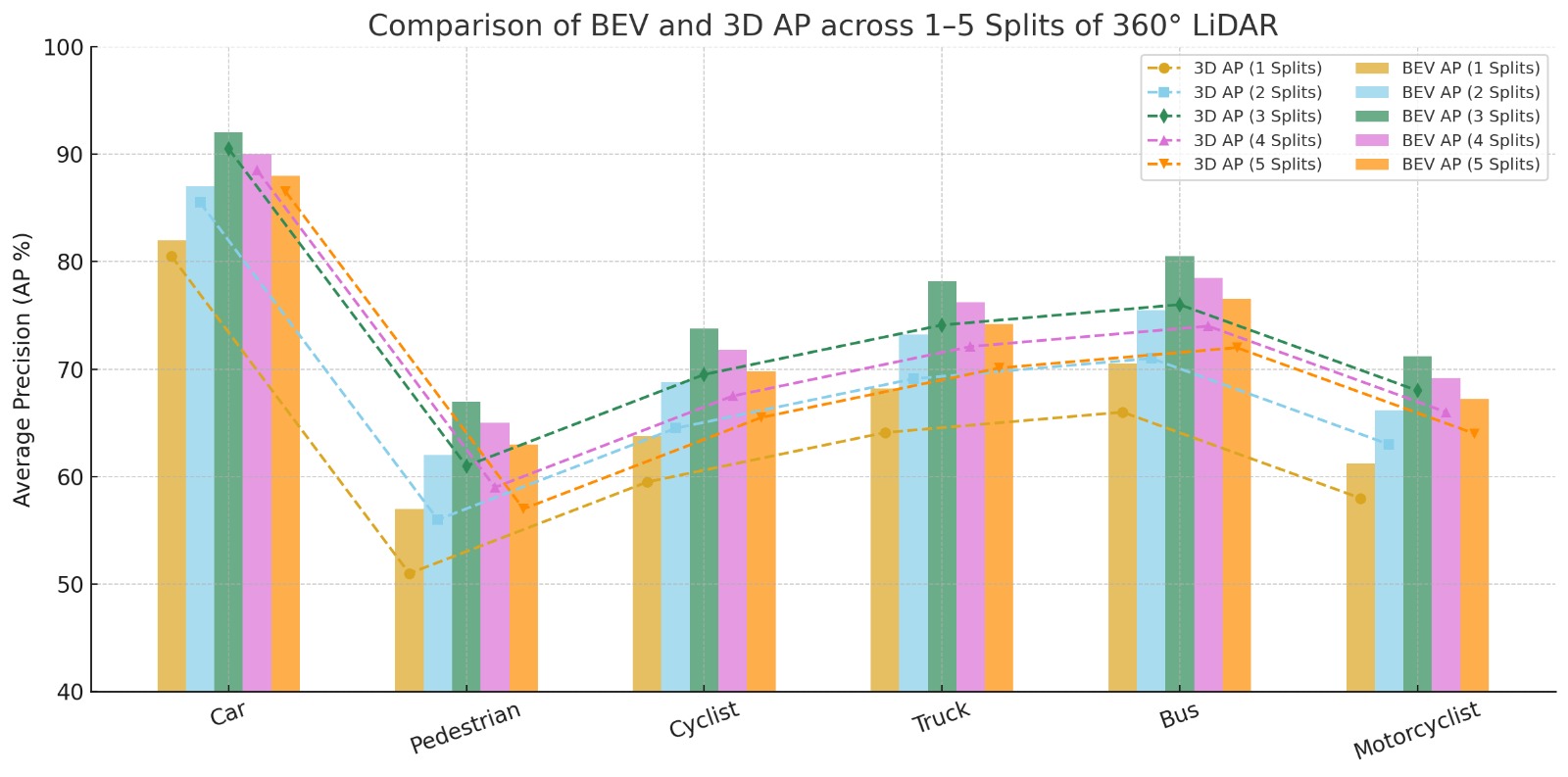}
\caption{BEV and 3D AP trends for different 360\textdegree\ sector split configurations.}
\label{fig:sectors}
\end{figure}

To understand what the azimuthal partitioning is doing, experiments were run with different numbers of angular sectors while keeping the rest of the pipeline unchanged. Figure~\ref{fig:sectors} summarizes how the sector count affects detection performance across these settings.

Using a single sector over the full 360\textdegree\ field of view tends to hurt performance for objects sitting close to angular discontinuities. Adding more sectors improves the handling of such boundaries but also introduces redundancy when the same object appears in multiple sectors. In the configurations tested here, three overlapping sectors turned out to be a reasonable compromise between coverage and stable behavior at sector boundaries for panoramic LiDAR scenes.

\subsection{Impact of Transformation-Equivariant Feature Extraction}
\label{subsec:equivariant_impact}

The role of transformation-equivariant feature extraction is examined for panoramic LiDAR sensing under a fixed sector layout. Instead of a strict, isolated ablation with and without equivariance, the focus here is on how detection behaves across changes in vehicle orientation and sensor heading.

Across the scenes considered, equivariant feature extraction is associated with more consistent localization of objects that appear at large azimuth angles or during maneuvers with substantial heading changes. This effect works alongside standard data augmentation and is particularly relevant in dense urban traffic, where viewpoint changes are frequent and sometimes abrupt.

These observations are meant to be read as qualitative evidence, since the work does not include a controlled numerical comparison that toggles equivariance on and off. The goal is to place equivariant modeling in context within a panoramic perception pipeline, rather than to claim standalone performance gains that can be attributed to equivariance alone.

\section{Conclusion}
\label{sec:conclusion}

This paper presented a 360\textdegree{} LiDAR-based perception pipeline for 3D object detection in dense, unstructured urban traffic, integrating sector-wise panoramic processing, adaptive voxelization, and transformation-equivariant feature extraction. Evaluation on a custom OS0 LiDAR dataset showed generally stable performance across multiple object categories under full-surround sensing. The detector performed best on larger and more structurally distinct classes, achieving 92.02/90.51 for cars, 80.53/76.34 for buses, and 78.59/74.16 for trucks. For smaller and more variable road users, the scores were lower, with 67.45/61.02 for pedestrians, 73.21/69.54 for cyclists, and 71.20/68.13 for motorcyclists, highlighting the continuing difficulty of detecting such classes in dense urban scenes.

These results suggest that combining omnidirectional LiDAR sensing with equivariance-aware feature learning can improve robustness to viewpoint variation and sector-boundary effects in challenging traffic environments. Rather than claiming direct superiority over methods evaluated on different datasets, this work shows that panoramic sensing and equivariant representations provide a useful basis for more reliable perception in unstructured road settings. The findings also underline that careful sector design and evaluation consistency become increasingly important when detection is extended to full 360\textdegree{} coverage. The current work leaves several directions open for future work. The evaluation is limited to daytime data collected under mostly fair-weather conditions, so the behavior of the pipeline under nighttime operation and adverse weather still needs to be examined more carefully. In addition, future work can explore richer temporal modeling and multi-modal fusion with complementary sensors such as cameras and radar to further improve robustness in real-world deployment.

\section*{Acknowledgements}
The authors would like to thank the Wipro--IISc Research and Innovation Network (WIRIN) Center of Excellence for Autonomous Vehicles, Bangalore, for access to data collection infrastructure and for technical support during the development of the dataset used in this work. The authors also acknowledge the efforts of colleagues and collaborators who contributed to data annotation and setup. Any opinions, findings, and conclusions expressed in this paper are those of the authors and do not necessarily reflect the views of the supporting organizations.

\bibliography{custom}
 
\end{document}